\documentclass[]{alaya}
\usepackage{makecell}
\usepackage{wrapfig}
\usepackage{tabularx}
\usepackage{textcomp}
\usepackage{stfloats}
\usepackage{url}
\usepackage{verbatim}
\usepackage{titlesec}
\usepackage{tocloft}
\usepackage{adjustbox}
\usepackage{multirow}
\usepackage{pifont}
\usepackage[sc]{mathpazo}
\IfFileExists{t1txtt.fd}{}{} 
\usepackage{tikz}
\usepackage{comment}
\usepackage{amsmath,amssymb}
\usepackage{colortbl}
\usepackage[numbers]{natbib}
\usepackage{color}
\usepackage{booktabs}
\usepackage{hyperref}
\usepackage{graphicx}
\usepackage{subcaption}
\RequirePackage{xspace}
\makeatletter
\DeclareRobustCommand\onedot{\futurelet\@let@token\@onedot}
\def\@onedot{\ifx\@let@token.\else.\null\fi\xspace}
\usepackage[most]{tcolorbox}
\usepackage{array}
\usepackage{siunitx}
\usepackage[table]{xcolor}
\usepackage{caption}
\definecolor{headerpurple}{HTML}{d8d2fc}
\definecolor{rowgray}{gray}{0.95}
\IfFileExists{CJKutf8.sty}{\usepackage{CJKutf8}}{}

\makeatother

\definecolor{adptorange}{RGB}{248, 205, 172}
\definecolor{cmpblue}{RGB}{189, 215, 238}

\definecolor{our_red}{RGB}{232,157,160}
\definecolor{our_blue}{RGB}{136,206,230}
\definecolor{our_orange}{RGB}{246,200,168}
\definecolor{our_green}{RGB}{178,211,164}

\definecolor{attn_code0}{RGB}{247,215,200}
\definecolor{attn_code1}{RGB}{238,169,139}
\definecolor{mlp_code0}{RGB}{204,201,221}
\definecolor{mlp_code1}{RGB}{102,95,153}
\definecolor{mygray}{HTML}{f0f0f0}

\definecolor{token_blue}{RGB}{84, 120, 140}

\IfFileExists{bbding.sty}{\usepackage{bbding}}{}
\IfFileExists{fontawesome.sty}{\usepackage{fontawesome}}{}
\usepackage{float}

\newlength\savewidth

\newcolumntype{x}[1]{>{\centering\arraybackslash}p{#1pt}}
\newcolumntype{y}[1]{>{\raggedright\arraybackslash}p{#1pt}}
\newcolumntype{z}[1]{>{\raggedleft\arraybackslash}p{#1pt}}

\renewcommand{\paragraph}[1]{\vspace{1.25mm}\noindent\textbf{#1}}

\usepackage{algorithm}
\usepackage{listings}

\definecolor{codeblue}{rgb}{0.25, 0.5, 0.5}
\definecolor{codekw}{rgb}{0.35, 0.35, 0.75}
\lstdefinestyle{Pytorch}{
  language = Python,
  backgroundcolor = \color{white},
  basicstyle = \fontsize{9pt}{8pt}\selectfont\ttfamily\bfseries,
  columns = fullflexible,
  aboveskip=1pt,
  belowskip=1pt,
  breaklines = true,
  captionpos = b,
  commentstyle = \color{codeblue},
  keywordstyle = \color{codekw},
}

\definecolor{green}{HTML}{009000}
\definecolor{red}{HTML}{ea4335}

\title{AnyGameRenderer: Entity-Consistent Generative Rendering Across Games and an Entity-Level Benchmark}
\author[1]{Ruicong Liu*} \author[1]{Liangyang Ouyang*} \author[1]{Fengbo Lan} \author[1]{Zhixiang Wang} \author[1,2]{Kaipeng Zhang}
\affiliation[1]{Alaya Lab} \affiliation[2]{Shanghai Innovation Institute}
\abstract{
  AI coding agents can now build small playable games in minutes, but the results usually look like programmer art: gray blockouts, colored geometric primitives, and flat semantic blobs. We present AnyGameRenderer, a video-to-video generative renderer that turns such footage into high-quality visuals with characters, motion detail, and effects. Throughout, it keeps the layout, camera, and every entity's position and timing intact.
  Our AnyGameRenderer generalizes across a wide range of genres that agents build, such as tower defense, auto-battlers, top-down survivors, turn-based battles, kart racing, and crowd runners.
  AnyGameRenderer takes plain video as its game-state input, with no complex representations such as engine G-buffers, per-object class labels, or low-poly proxies. Gray blockouts, colored primitives, semantic maps, and wireframes all work as input.
  It also preserves entity correspondence: group counts stay correct, and entities appear and disappear on the right frame.
  We implement AnyGameRenderer by LoRA fine-tuning MiniMax-H3 ref2va, a 33B reference-video-conditioned video generation model. Training uses $\sim$10k paired clips from three sources: (i) multi-stream rendering data from 38 agent-made games; (ii) two-stream rendering data from Unreal Engine, with programmatically controlled crowds, count changes, and entity scales; and (iii) real gameplay recordings, with pairs reconstructed via 3D-box detection.
  To evaluate the entity correspondence, we propose an entity-level benchmark of 59 held-out cases with 753 entities, scored against ground-truth boxes. Compared with the base model, AnyGameRenderer better preserves entity correspondence: temporal match rises from 50.9\% to 62.9\%, missed count changes fall from 34.7\% to 21.3\%, and fewer entities are hallucinated or left unrendered.
}

\github{\url{https://alaya-lab.github.io/AnyGameRenderer}}
\Code{\url{https://github.com/AlayaLab/AnyGameRenderer}}
\correspondence{Kaipeng Zhang}
\note{* Equal contribution}
\date{\today}

\begin{document}
\maketitle

\section{Introduction}
\label{sec:intro}

Coding agents and game-creation tools can now turn a one-paragraph prompt into a playable game in minutes, complete with game logic, physics, camera, win conditions, and working HUD~\cite{chen2026recursivegamecreator}. What they cannot produce is good art. The output usually contains gray boxes, colored cylinders, and cones on flat-shaded ground. Human studios spend months replacing this kind of ``programmer art'' through modeling, texturing, and lighting.
Splitting a game into an engine that manages state and a generative model that renders pixels has quickly become a common idea. Concurrent systems such as CWM~\cite{chen2026cwm}, PWM~\cite{huang2026pwm}, Matrix-Game~4.0~\cite{skywork2026matrixgame4}, StreamRender-H3~\cite{nvidia2026streamrenderh3}, Marionette \cite{meng2026marionette}, and SolarRenderer~\cite{solarrenderer2026} all adopt it in some form. We do not argue for this setting. Instead, we build a rendering layer that is more \emph{general} in what it accepts and more \emph{faithful} in what it keeps.

\begin{figure}[t!]
  \centering
  \includegraphics[width=.9\linewidth]{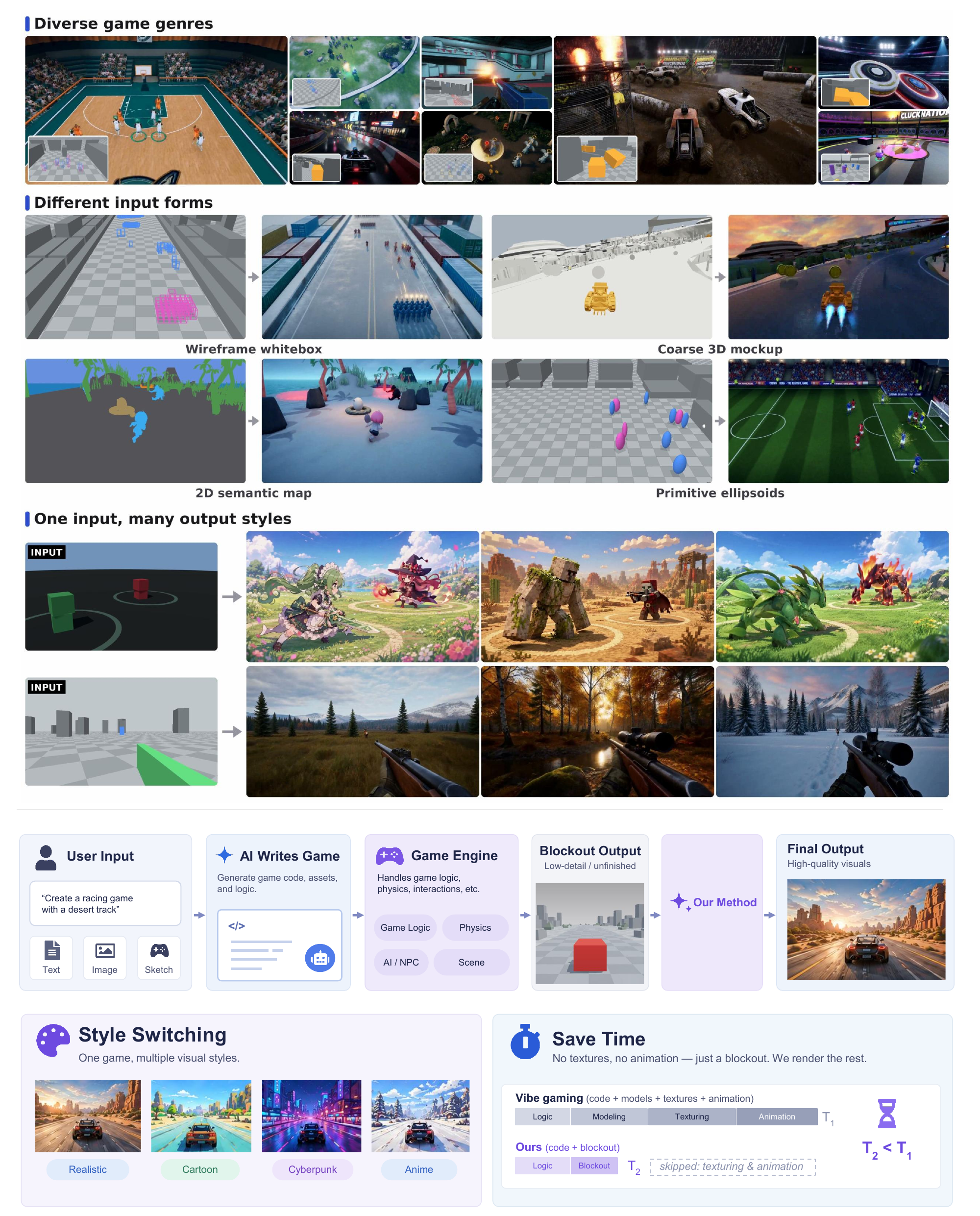}\vspace{2pt}
\caption{\textbf{AnyGameRenderer turns crude gameplay into polished game visuals.}
\emph{Top:} one checkpoint handles diverse genres and input types, and renders one input in many styles. Insets show the crude input.
\emph{Bottom:} an agent writes the game and the engine outputs a blockout, which AnyGameRenderer renders into the final visuals. This allows style switching and skips modeling, texturing and animation.}
\label{fig:teaser}
\end{figure}

As shown in \Cref{fig:teaser}, AnyGameRenderer takes a crude gameplay recording and a text prompt (optionally with a style anchor image), and outputs the same gameplay re-rendered in the prompted style, frame-aligned with the input. It adds characters, motion detail, and effects, while keeping the layout, camera, and every entity's position and timing. This brings two practical benefits. A game can easily switch to a new visual style by changing only the prompt. It also makes vibe gaming faster, since a game only needs code and a blockout, without modeling, texturing or animation.

There are three properties that set AnyGameRenderer apart.
\emph{(1) It generalizes across genres.} Prior and concurrent works are demonstrated mainly on third-person, first-person, or driving footage, with limited coverage of the genres that agents actually produce. As illustrated in \Cref{fig:teaser}, AnyGameRenderer generalizes across a wide range of genres, including tower defense, auto-battlers, top-down survivors, turn-based battles, kart racing, crowd runners, arena survival, co-op games, and shooters. In addition, it accepts diverse input types, including gray blockouts, colored primitives, semantic maps, wireframes, and coarse mockup.

\emph{(2) It keeps entity correspondence.} The hard part of game rendering is bookkeeping: twelve enemies become nine when three die; a reinforcement wave walks in at $t{=}4.2$\,s; a tiny far-away mob must not dissolve into fog. AnyGameRenderer substantially improves entity correspondence: group counts stay correct, entities appear and disappear on the right frame, and fewer small entities are lost. The zero-shot base model often fails here: entities never appear, colored blocks stay unrendered, and characters appear even when they are not in the input.

\emph{(3) Its input is plain video.} Many concurrent systems rely on costly or complex representations, such as per-object class labels (StreamRender-H3, CWM), state-augmented 3D boxes (PWM), or low-poly engine proxies (Matrix-Game~4.0). AnyGameRenderer instead accepts whatever a crude game looks like, including gray blockouts, colored primitives, semantic-colored maps, wireframes, and coarse clay figures, at different granularities and with one checkpoint.

To achieve these properties, we build paired (crude video, styled video) data from three sources. (i)~Multi-stream rendering data from 38 agent-made games helps the model generalize across genres and input types. (ii)~Two-stream rendering in Unreal Engine uses scenes scripted with same-color groups that change count mid-clip, entity sizes from 20 to 600\,px, and frame-aligned appearance and disappearance, teaching the model to keep entity correspondence. (iii)~Real gameplay recordings, converted into crude inputs by 3D-box detection, help the model render realistic scenes. Together they yield 10,330 training clips.

Existing renderers are evaluated at the frame level, with edge agreement, point tracks, or pairwise judges, which cannot tell whether each entity is rendered correctly. We therefore design an entity-level benchmark of 59 held-out cases with 753 entities, each annotated with ground-truth boxes, and evaluate it with four metrics. \emph{Temporal match} measures whether each entity appears and disappears on time; \emph{position match} measures whether it stays at the right place; \emph{unrendered rate} measures how often crude input blocks are left unrendered; and \emph{hallucination rate} measures how often the output contains entities that do not exist in the input. Compared with the zero-shot base, AnyGameRenderer raises temporal match from 50.9\% to 62.9\% and position match from 57.1\% to 61.9\%, and lowers the unrendered rate from 4.1\% to 2.9\% and the hallucination rate from 52.1\% to 46.0\%.

\paragraph{Contributions.}
\begin{itemize}
    \setlength\itemsep{1pt}
  \item We propose a video-to-video generative renderer for games. A single checkpoint handles a wide range of genres and input types, renders them in diverse prompted styles, and preserves entity correspondence with the input.
  \item We build a paired dataset with three sources: multi-stream rendering data from agent-made games, two-stream rendering data from Unreal Engine with scenarios designed for correspondence, and detection-based reconstruction from real gameplay recordings. Together they yield 10,330 training clips.
  \item We propose an entity-level benchmark and metrics for entity correspondence. The metrics measure whether each entity appears on time, stays in place, and is neither unrendered nor hallucinated. On this benchmark, AnyGameRenderer improves temporal match from 50.9\% to 62.9\% and reduces missed count changes from 34.7\% to 21.3\% over the zero-shot base.
\end{itemize}

\section{Related Work}
\label{sec:related}

\subsection{AI-driven game creation}
\label{sec:related_game_creation}

Early work uses large language models to generate game content such as levels and rules. MarioGPT~\cite{sudhakaran2023mariogpt} generates Super Mario Bros.\ levels from text, Word2World~\cite{nasir2024word2world} turns stories into playable 2D tile worlds, and LLMGG~\cite{hu2024game} generates game rules and levels in the Video Game Description Language. Multi-agent frameworks such as GameGPT~\cite{chen2023gamegpt} split game development into roles that collaborate on a full project. More recently, coding agents build complete games end-to-end. Recursive Game Creator~\cite{chen2026recursivegamecreator} iterates designer, builder, player and reviewer roles over HTML and Godot projects; OpenGame~\cite{jiang2026opengame} is an open-source agentic framework for web games across genres such as action and tower defense. While these systems produce working game logic, their visuals are usually limited to simple shapes and placeholder assets, far from the art quality of commercial games.

\subsection{Interactive world models}
\label{sec:related_world_models}

Another line of work replaces the game engine altogether: an autoregressive video model, conditioned on player actions, generates the next frames. Examples include Genie~\cite{bruce2024genie,parker2024genie,ball2025genie3}, GameNGen~\cite{valevski2025diffusion}, Oasis~\cite{decart2024oasis}, MineWorld~\cite{guo2025mineworld}, GameGen-X~\cite{che2025gamegen}, GameFactory~\cite{yu2025gamefactory}, Hunyuan-GameCraft~\cite{li2025hunyuan,tang2025hunyuan}, the Matrix-Game line~\cite{he2025matrix,wang2026matrix,feng2026matrix}, and Yume~\cite{mao2025yume,Mao_2026_CVPR}. These models are strong simulators but hard to control in the ways a game needs: exact entity counts, scripted events, and rules that hold for minutes. Long-horizon memory~\cite{xiao2025worldmem,yu2025contextasmemory,wu2026infiniteworld,liu2025worldweaver} reduces drift but does not make the world programmable.

\subsection{Generative rendering for games}
\label{sec:related_rendering}

A growing body of work keeps the engine in charge of game state and uses a generative model only to render pixels. Neural forward rendering from G-buffers~\cite{liang2025diffusion,beisswenger2025framediffuser,zhang2026renderflow,lu2026germ} and earlier generative world renderers~\cite{huang2026generative,lin2026alayarendererflash} follow this split, but require engine-exact inputs. Several concurrent systems extend it to \emph{crude} inputs that a game or an agent can produce cheaply.
CWM~\cite{chen2026cwm} lets a coding agent maintain world state and renders a 12-class proxy video with a MiniMax-H3 LoRA. PWM~\cite{huang2026pwm} converts real AAA recordings into state-augmented 3D oriented boxes and renders them with a structured spatial ControlNet on a frozen backbone. Matrix-Game~4.0~\cite{skywork2026matrixgame4} renders low-poly engine proxies in real time with a distilled causal DiT. StreamRender-H3~\cite{nvidia2026streamrenderh3} streams a 13-class semantic map into video. SolarRenderer~\cite{solarrenderer2026} trains a control LoRA on a frozen MiniMax-H3 with six types of structural video, passes them without type labels, and generalizes to structure types unseen in training; it also renders in four steps and supports streaming.

AnyGameRenderer differs from these systems in three ways. First, they are demonstrated mainly on realistic third-person, driving, or a few self-built 3D scenes; our method generalizes across a wide range of genres, including tower defense, auto-battlers, top-down survivors, turn-based battles, kart racing, crowd runners, arena survival, co-op games, and shooters. Second, many of them rely on a fixed intermediate representation, while we take plain video. Third, none of them measures entity-level correspondence. Frame-level metrics barely change when a small entity is missing or a group arrives late, and edge agreement even rewards leaving crude blocks unrendered. We make correspondence, including the count, timing and position of every entity, the main target of both training data and evaluation.

\section{Method}
\label{sec:method}

\begin{figure}[t]
  \centering
  \includegraphics[width=\linewidth]{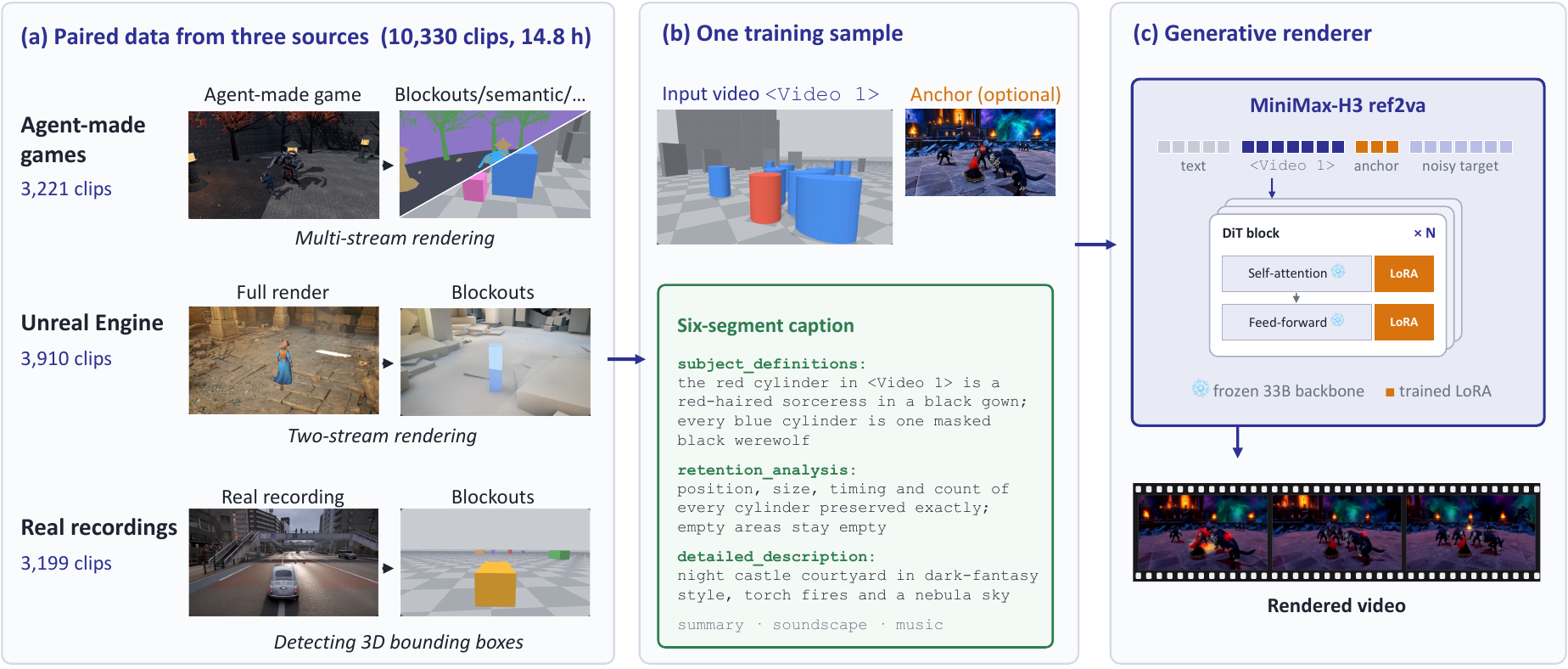}
  \caption{\textbf{Overview of AnyGameRenderer.}
    (a)~Paired data from three sources. Agent-made games use multi-stream rendering, which pairs each game render with several aligned crude streams such as blockouts and semantic maps. Unreal Engine scenes use two-stream rendering, which pairs a full render with blockouts. Real gameplay recordings are converted into blockouts by detecting 3D bounding boxes.
    (b)~One training sample: the crude clip is the input video \texttt{<Video 1>}, the target's first frame serves as an optional anchor, and a six-segment caption defines which real entity each crude shape represents and what must be preserved.
  (c)~Text, \texttt{<Video 1>}, anchor, and noisy target tokens form one sequence for the frozen MiniMax-H3 DiT. Only the LoRA adapters on self-attention and feed-forward layers are trained, and the model renders the input gameplay in the prompted style.}
  \label{fig:method}
\end{figure}

AnyGameRenderer treats game rendering as reference-conditioned video editing: given a crude gameplay recording (a plain reference video) and a caption saying what each crude entity stands for, it generates the same gameplay in a target style (\Cref{fig:method}). We design a caption schema to enforce entity correspondence (\Cref{sec:caption}).

\subsection{Task setting}
\label{sec:task}

\paragraph{Task.}
Let $V_{\mathrm{in}}\in\mathbb{R}^{T\times H\times W\times 3}$ be a crude gameplay clip and $c$ be a text prompt describing the target style and what each crude entity represents. AnyGameRenderer produces $V_{\mathrm{out}}$ of the same length and resolution such that (i)~camera, layout, and timing match $V_{\mathrm{in}}$ frame by frame; (ii)~every crude entity becomes its finished counterpart at the same position, apparent size, and heading, appearing and disappearing on the same frames; and (iii)~appearance, including materials, lighting, character design, and environment, follows $c$ and an optional anchor image $I_{\mathrm{a}}$.

\paragraph{Base model.}
We build on the ref2va mode of MiniMax-H3~\cite{minimax2026h3}, which edits a reference video according to a text prompt. We pass the crude clip as \texttt{<Video 1>} and describe the target rendering in the prompt, so any crude input works. The model runs at 24\,fps up to 768p. We disable the audio loss in training and discard audio at inference.

\paragraph{Conditioning.}
Each sample has three conditions: 1) the crude clip as reference video \texttt{<Video 1>}; 2) a six-segment structured caption (\Cref{sec:caption}); and 3) an optional anchor image. During training, the anchor is the target's first frame; at inference, it serves as a style reference or is omitted. The crude clip fixes the structure, while the caption and anchor set the style. To render the same clip in a new style, we only change the prompt or the anchor.

\paragraph{Trainable parameters.}
We train a LoRA~\cite{hu2022lora} of rank 128 on all transformer blocks with the standard flow-matching objective~\cite{lipman2023flow}, keeping the base weights frozen. All reported results use 20-step Euler sampling on base$+$LoRA.

\subsection{Caption schema: the correspondence contract}
\label{sec:caption}

We keep the base model's six-segment format (\texttt{subject\_definitions}, \texttt{summary}, \texttt{retention\_analysis}, \texttt{detailed\_description}, \texttt{overall\_soundscape}, \texttt{non\_diegetic\_music}) and follow four rules, which we call the correspondence contract.

\paragraph{Subjects are entities; crude shapes are their representation.}
Each subject is defined as the real entity, and the crude input is named as its representation, e.g., ``\emph{every blue ellipsoid in} \texttt{<Video 1>} \emph{is one raccoon-like orchard raider, and the whole blue group is a group of identical ones}.'' The color-to-identity mapping is set per clip, groups are described at class level with explicit statements that they exist.

\paragraph{Retention analysis fixes what follows the input.}
This segment declares position, apparent size, timing, and head count as preserved exactly, and leaves appearance to the prompt. It also describes empty space positively: ``areas empty in \texttt{<Video 1>} stay empty; no character is added that has no primitive in \texttt{<Video 1>}.''

\paragraph{Positive assertions only.}
The base model has no negative prompt, and negative phrasing tends to produce what it names (``no HUD'' reliably produces a HUD). All captions are therefore written as positive statements.

\paragraph{Programmatic captions with deliberate noise.}
Captions are generated automatically from engine metadata and detection results, not written by hand. Each caption lists every character that may appear in the game, even if some never show up in the clip. We keep this noise on purpose: it teaches the model to decide what exists from the video, not from the text. As a result, the model rarely adds characters that are named in the caption but missing from the input.

\subsection{Training recipe}
\label{sec:recipe}

\paragraph{Unified format.}
All sources are resampled to 124-frame clips at 24\,fps (5.17\,s) and $1344{\times}768$, with the target's first frame as the anchor and a silent 32\,kHz audio track. Latents and text embeddings are cached once per source.

\paragraph{Optimization.}
We train a rank-128 LoRA on all blocks with a global batch size of 16. The learning rate warms up over 100 steps and then follows a cosine decay, with a peak between $2{\times}10^{-5}$ and $2.8{\times}10^{-5}$; weight decay is 0.01. We use bf16 precision, gradient checkpointing, and block swapping, while FlashAttention-3~\cite{shah2024flashattention3} speeds up each step by about 30\%. Training runs on 16 NVIDIA H200 GPUs (141\,GB). Each step takes about 129\,s, and three epochs over all 10,330 training clips take about 48 hours.


\paragraph{Deployment.}
We use a LoRA multiplier of 1.0 for in-domain inputs and styles. For out-of-domain text-only styles (no anchor image, and a style unseen in training), a multiplier of 0.7 recovers the zero-shot visual quality while keeping the correspondence gains. We recommend it as the default for open-ended styling.

\section{Data}
\label{sec:data}

We build (crude video, styled video) pairs from three sources: multi-stream rendering data from agent-made games covers different genres and input types; two-stream rendering data from Unreal Engine targets entity correspondence; and real gameplay recordings add realistic layouts, cameras, and motion. In every pair, the crude and styled videos share the same camera and layout and are frame-aligned by construction.

\subsection{Agent-made games}
\label{sec:agent_games}

Our multi-stream rendering data comprises 38 agent-made games, covering shooters, survivors, tower defense, kart racing, arena, and co-op designs. This source contributes \textbf{3,221} clips. Each clip has five aligned passes: the target render; a primitive pass, where the scene is a gray blockout and characters are simple geometric shapes; a coarse pass, with a more detailed blockout and low-poly clay figures; a 12-class semantic map; and depth. Each clip also comes with an anchor image, a caption, and 3D boxes.
We train on the pair of primitive and target.

\subsection{Unreal Engine}
\label{sec:ue}

Each Unreal Engine scene uses two-stream rendering: it is rendered from the same simulation and camera as a gray blockout with colored blocks for characters and as a full render. The protagonist is always blue, and other characters use a six-color set.

\begin{figure}[t]
  \centering
  \includegraphics[width=\linewidth]{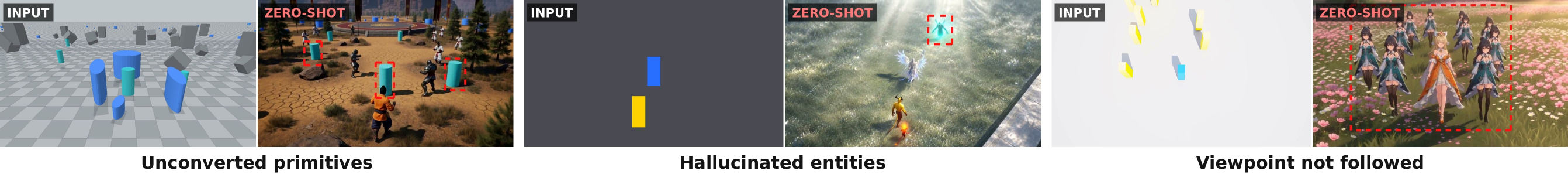}
  \caption{\textbf{Three failures of the zero-shot base model.} Red dashed boxes mark the errors.
    \emph{Left:} input primitives are left unrendered; the cyan cylinders stay in the output instead of becoming characters.
    \emph{Middle:} the input has only two blocks, but the model adds an extra creature that does not exist in the input.
  \emph{Right:} the input is a top-down view, but the model renders a frontal group shot.}
  \label{fig:motivation}
\end{figure}

Rather than collecting generic gameplay, we script scenes around the zero-shot failures (see examples in \Cref{fig:motivation}) to ensure entity correspondence:
\begin{itemize}
    \setlength\itemsep{1pt}
  \item \textbf{Count changes in same-color groups.} Groups of 2 to 20 characters walk in, spawn, walk out, or despawn during the clip, in dense clusters, columns, or scattered formations. This targets crowds left as unrendered blocks and entities that appear late or never.
  \item \textbf{Characters of different colors entering and leaving.} This targets wrong entry and exit timing when several character types share a scene.
  \item \textbf{Entity sizes from 20 to 600 pixels}, including boss-sized characters and characters moving toward or away from the camera. This targets small, distant entities that dissolve into the background.
  \item \textbf{Mixed camera views}: 35\% follow, 30\% top-down, 20\% isometric, and 15\% side. This keeps the model from relying on a single viewpoint.
\end{itemize}
We keep clips where the protagonist is visible in at least 95\% of frames. Each clip carries metadata on the scene, characters, colors, and events, from which its caption is generated. This source contributes \textbf{3,910} clips.

\subsection{Real gameplay recordings}
\label{sec:pwm}

Following the reconstruction approach in PWM~\cite{huang2026pwm}, we convert real gameplay recordings of combat, driving, and pedestrian scenes into crude inputs. We detect 3D bounding boxes in each frame and render them over a checkerboard ground as proxies of varied shapes: solid boxes, wireframe boxes, cylinders, and upright ellipsoids, with about 800 clips per shape. The original recording is the target, and captions come from the detection labels plus one sentence describing the environment. This source adds real-world scenes, with realistic camera work and vehicle and pedestrian motion. It contributes \textbf{3,199} clips.

The three sources are mixed at roughly equal weight (\Cref{tab:mixture}), giving 10,330 clips of 124 frames, about 14.8 hours of paired video.

\begin{table}[t]
  \centering
  \caption{Training data mixture. All clips are 124 frames at 24\,fps and $1344{\times}768$, with captions generated from each source's metadata.}
  \small
  \begin{tabularx}{\linewidth}{@{}l >{\raggedright\arraybackslash}X r r@{}}
    \toprule
    Source & Crude input & Clips & Share \\
    \midrule
    Agent-made games (38) & primitive / coarse / semantic passes & 3,221 & 31\% \\
    Unreal Engine & gray blockout with colored character blocks & 3,910 & 38\% \\
    Real recordings & detected 3D boxes as solid / wireframe / cylinder / ellipsoid proxies & 3,199 & 31\% \\
    \midrule
    \textbf{Total} & & \textbf{10,330} & $\approx$14.8\,h \\
    \bottomrule
  \end{tabularx}
  \label{tab:mixture}
\end{table}

\begin{table}[t]
  \centering
  \caption{\textbf{Entity-sync results (\%) on the held-out benchmark}: four main metrics (left) and three secondary statistics (right). \emph{GT floor} (gray) runs the ground-truth RGB video through the same detection and matching pipeline.}
  \small
  \setlength{\tabcolsep}{2.5pt}
  \newcommand{\gt}[1]{\textcolor{gray}{#1}}
  \begin{tabular}{llcccc|ccc}
    \toprule
    & & Temporal & Position & Unrendered & Halluc. & On-time & Never & Missed count \\
    Split & Model & match $\uparrow$ & match $\uparrow$ & rate $\downarrow$ & rate $\downarrow$ & appear. $\uparrow$ & appears $\downarrow$ & changes $\downarrow$ \\
    \midrule
    \multirow{3}{*}{All (59)}
    & Zero-shot & 50.9 & 57.1 & 4.1 & 52.1 & 55.9 & 26.0 & 34.7 \\
    & \textbf{Ours} & \textbf{62.9} & \textbf{61.9} & \textbf{2.9} & \textbf{46.0} & \textbf{67.6} & \textbf{20.2} & \textbf{21.3} \\
    & \gt{GT floor} & \gt{80.9} & \gt{79.3} & \gt{0.1} & \gt{34.1} & \gt{93.6} & \gt{0.7} & \gt{0.0} \\
    \midrule
    \multirow{3}{*}{Multi-stream rendering (34)}
    & Zero-shot & 47.6 & 56.4 & 4.5 & 55.4 & 50.9 & 28.4 & 40.7 \\
    & \textbf{Ours} & \textbf{60.7} & \textbf{60.6} & \textbf{3.2} & \textbf{51.2} & \textbf{64.0} & \textbf{23.5} & \textbf{23.7} \\
    & \gt{GT floor} & \gt{79.3} & \gt{76.8} & \gt{0.1} & \gt{35.7} & \gt{91.5} & \gt{0.9} & \gt{0.0} \\
    \midrule
    \multirow{3}{*}{Two-stream rendering (11)}
    & Zero-shot & 80.2 & 68.1 & 4.9 & 38.9 & 82.3 & 15.2 & 0.0 \\
    & \textbf{Ours} & \textbf{96.6} & \textbf{80.6} & \textbf{1.4} & \textbf{19.4} & \textbf{94.9} & \textbf{0.0} & \textbf{0.0} \\
    & \gt{GT floor} & \gt{98.5} & \gt{86.7} & \gt{0.3} & \gt{16.6} & \gt{100.0} & \gt{0.0} & \gt{0.0} \\
    \midrule
    \multirow{3}{*}{Real game (14)}
    & Zero-shot & 46.5 & 52.4 & \textbf{2.6} & 52.1 & 61.3 & 22.6 & 20.0 \\
    & \textbf{Ours} & \textbf{51.1} & \textbf{53.2} & 2.9 & \textbf{45.4} & \textbf{66.1} & \textbf{18.5} & \textbf{20.0} \\
    & \gt{GT floor} & \gt{76.7} & \gt{85.5} & \gt{0.0} & \gt{37.4} & \gt{99.2} & \gt{0.0} & \gt{0.0} \\
    \bottomrule
  \end{tabular}
  \label{tab:main}
\end{table}

\section{Experiments}
\label{sec:exp}

We first describe the evaluation protocol (\Cref{sec:protocol}), then evaluate entity correspondence (\Cref{sec:main}) and generalization across genres (\Cref{sec:genre}).

\subsection{Evaluation protocol}
\label{sec:protocol}

\paragraph{Benchmark.}
The benchmark contains 59 cases with 753 annotated entities: 11 held-out two-stream rendering cases from Unreal Engine (gray blockouts), 14 real recordings with detected solid boxes, and 34 multi-stream rendering cases from six held-out agent-made games. Each case uses a source clip excluded from training, rendered in a style or with characters not seen in training. All models are run with the same seed, 20 sampling steps, and the same captions.

\paragraph{Entity-sync metrics.}
Input entities are the ground-truth 2D boxes shipped with each crude video; entities that exist for fewer than 12 frames are ignored. In the generated video, we locate characters with SAM3 open-vocabulary segmentation, prompted with the caption's subject phrases plus generic person and vehicle terms. We remove shadow and reflection detections and link the remaining detections into tracks. In each frame, input boxes and detections are matched with the Hungarian method, and a pair is accepted only if IoU${}\ge 0.3$ or the centroid offset is within half the input box size. We report four main metrics, averaged over entities:
\begin{itemize}
    \setlength\itemsep{1pt}
  \item \textbf{Temporal match}: the IoU between the frames in which an entity exists in the input and the frames in which its track is in place. Late or missing appearance, lingering after despawn, and identity swaps all lower it.
  \item \textbf{Position match}: $1-d/(0.5\,s)$ on matched frames, where $d$ is the centroid offset and $s$ is the input box size.
  \item \textbf{Unrendered rate}: the fraction of entity-frames in which the input block is still visible as a flat color patch.
  \item \textbf{Hallucination rate}: the fraction of detected entity-frames that belong to tracks matching no input entity.
\end{itemize}
We also report three secondary statistics:
\begin{itemize}
    \setlength\itemsep{1pt}
  \item \textbf{On-time appearance}: an entity is recognized at the first frame $t$ where it is matched in both $t$ and $t{+}1$, giving an onset delay $\Delta_{\mathrm{on}} = t - t_{\mathrm{in}}$, where $t_{\mathrm{in}}$ is its first frame in the input. An entity is on time if $|\Delta_{\mathrm{on}}| \le 4$ frames (${\approx}0.17$\,s).
  \item \textbf{Never appears}: the fraction of entities that are never recognized.
  \item \textbf{Missed count changes}: a count change is a rise of at least two entities in the input within six frames, $n_{\mathrm{in}}[t{+}3]-n_{\mathrm{in}}[t{-}3]\ge 2$, with target count $n_{\mathrm{in}}[t{+}8]$. It is reproduced if, within the next 60 frames, the detected count stays at or above the target minus one for three consecutive frames. We report the fraction of the 75 count changes in the benchmark that are missed.
\end{itemize}
The \emph{GT floor} in \Cref{tab:main} runs the ground-truth RGB video through the same pipeline and marks the best score the pipeline can measure.

\subsection{Entity correspondence}
\label{sec:main}

\paragraph{Quantitative results.}
\Cref{tab:main} reports results on all 59 cases and on each source. Over all cases, our method improves all four main metrics: temporal match rises from 50.9\% to 62.9\%, position match from 57.1\% to 61.9\%, the unrendered rate falls from 4.1\% to 2.9\%, and the hallucination rate falls from 52.1\% to 46.0\%. The secondary statistics show the same trend. On-time appearance rises from 55.9\% to 67.6\%, entities that never appear fall from 26.0\% to 20.2\%, and missed count changes fall from 34.7\% to 21.3\%.

Gains differ by source. On the multi-stream rendering cases from agent-made games, temporal match rises by 13 points and missed count changes nearly halve. On Unreal Engine's two-stream rendering cases, temporal match reaches 96.6\%, close to the GT floor of 98.5\%, and no entity is missed. Real recordings improve the least: temporal match rises from 46.5\% to 51.1\%, the unrendered rate is similar, and missed count changes stay at 20.0\%.

\begin{figure}[t]
  \centering
  \includegraphics[width=\linewidth]{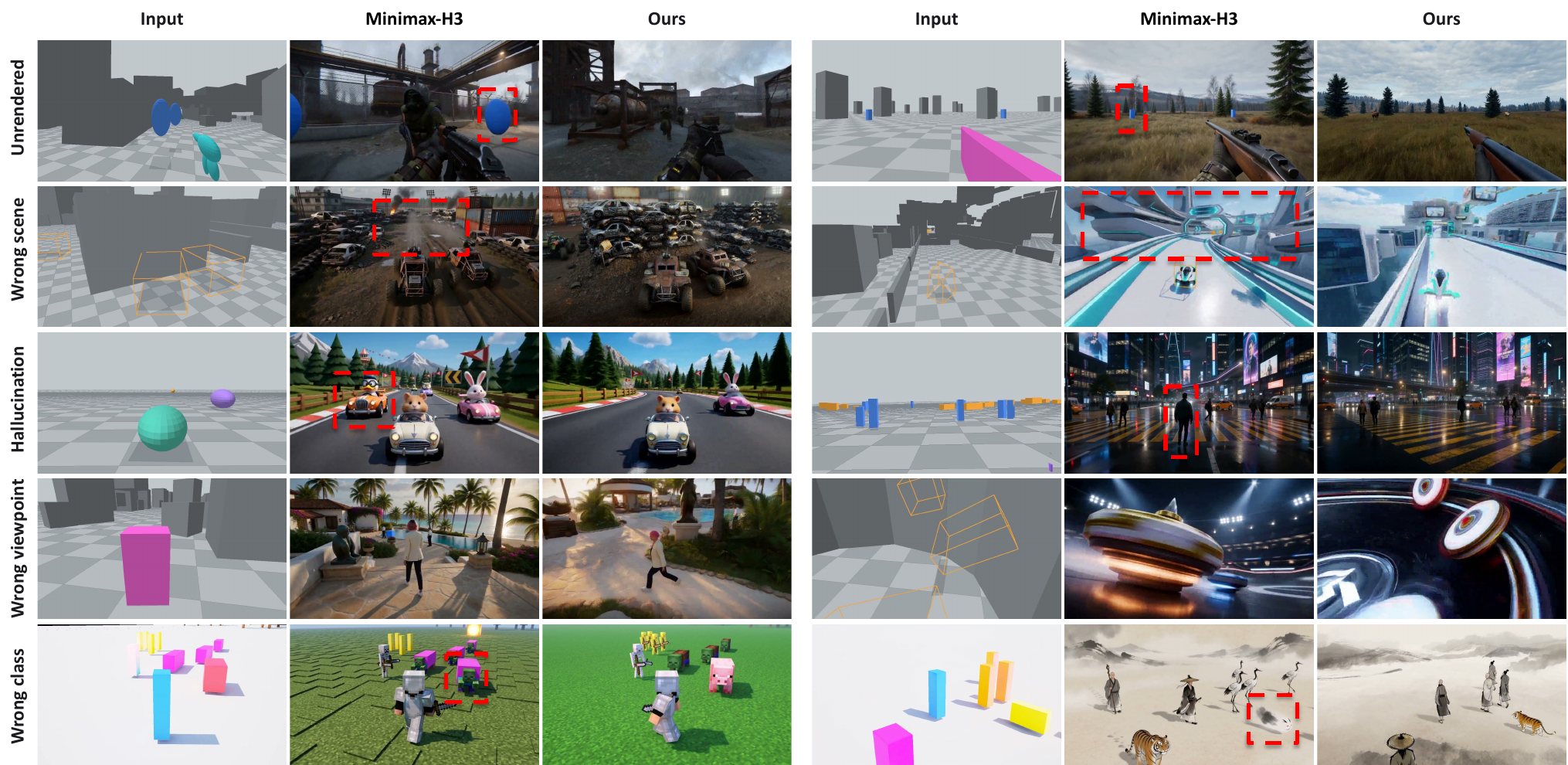}
  \caption{\textbf{Zero-shot MiniMax-H3 versus AnyGameRenderer on held-out cases}, with the same seed, steps, and caption. The zero-shot model leaves crude shapes unrendered (rows 1--4), drops the input viewpoint for out-of-domain styles (rows 5--6: a top-down blockout becomes a frontal anime scene or a first-person voxel scene), and adds caption-named characters although the input has only two blocks (row 7). AnyGameRenderer fixes all three. For rows 5--6, the ``GT'' column is the original in-domain render and serves only as a layout reference.}
  \label{fig:compare}
\end{figure}

\paragraph{Qualitative comparison.}
\Cref{fig:compare} shows seven held-out cases. The zero-shot model works well on sparse scenes with clean captions, but on game footage it fails in three ways. First, it leaves crude shapes in the output: blue cubes in an orchard, cyan cylinders in an arena, and a blue cylinder in the grass where an elk should be. Second, it drops the input viewpoint when the style is far from typical game footage, turning a top-down blockout into a frontal anime group shot or a first-person Minecraft-like scene. Third, it adds characters named in the caption that have no block in the input. AnyGameRenderer fixes all three in these cases while keeping the prompted style.

\subsection{Generalization across genres}
\label{sec:genre}

\begin{figure}[t]
  \centering
  \includegraphics[width=\linewidth]{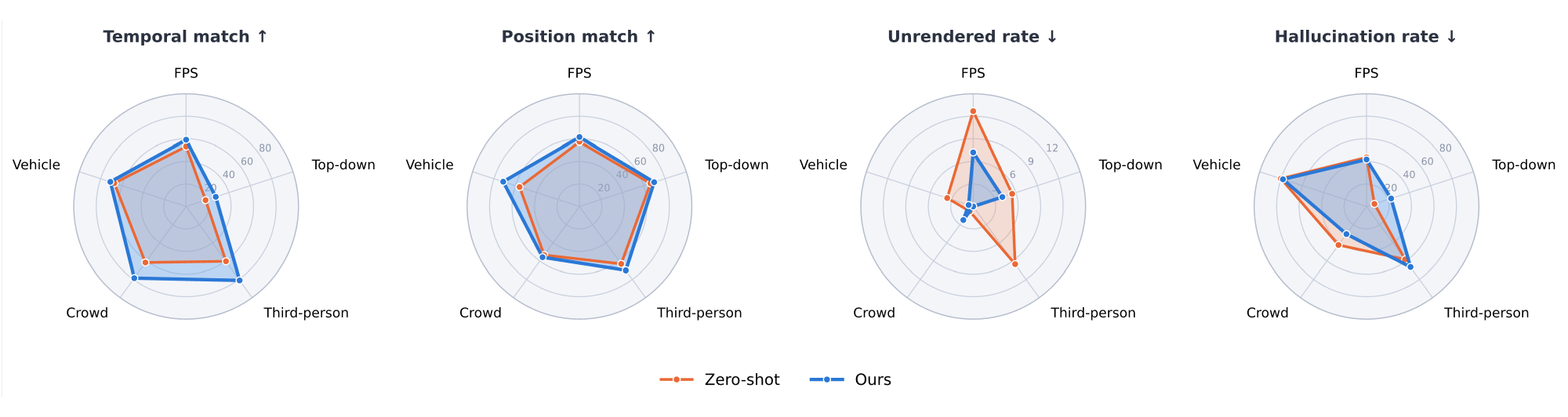}
  \caption{\textbf{Entity-sync metrics by genre.} Zero-shot MiniMax-H3 (orange) versus AnyGameRenderer (blue) on the held-out benchmark, grouped into first-person shooter, top-down, third-person, crowd, and vehicle games.}
  \label{fig:genre}
\end{figure}

A single checkpoint handles all five genres in our benchmark: first-person shooters, top-down, third-person, crowd, and vehicle games (\Cref{fig:genre}). AnyGameRenderer improves temporal match over the zero-shot model in every genre, with the largest gains on crowd and third-person games. It also lowers the unrendered rate in every genre except crowd games, where both models are already near zero, and improves position match most on vehicle games. The hallucination rate falls on crowd games but rises slightly on top-down and third-person games. Top-down games remain the hardest, with low temporal match for both models.

\section{Conclusion}
\label{sec:conclusion}

We presented AnyGameRenderer, a video-to-video generative renderer that turns the programmer art of agent-made games into high-quality visuals. We train a single LoRA on MiniMax-H3, using paired data from agent-made games, Unreal Engine's two-stream rendering, and real gameplay recordings. The resulting checkpoint handles a wide range of genres. It takes plain video as input, from gray blockouts and colored primitives to semantic maps and wireframes. It also keeps entities in sync with the input. On a held-out benchmark, temporal match rises from 50.9\% to 62.9\%, and missed count changes fall from 34.7\% to 21.3\%. As coding agents build more and more games, we believe a general renderer that keeps every entity in place will help these games look as good as they play.

\section{Acknowledgments}
\label{sec:ack}

We thank Zheng-Hui Huang for providing the real gameplay recordings and the detection method used to process it. We also thank Xiaojie Xu and Zhengyuan Lin for providing the Unreal Engine rendering repository.

\bibliographystyle{abbrv}
\bibliography{references}

\end{document}